\documentclass{article}
\usepackage{kr}

\usepackage{times}
\usepackage{soul}
\usepackage{url}
\usepackage[hidelinks]{hyperref}
\usepackage[utf8]{inputenc}
\usepackage[small]{caption}
\usepackage{graphicx}
\usepackage{amsmath}
\usepackage{amsthm}
\usepackage{booktabs}
\usepackage{algorithm}
\usepackage{algorithmic}
\usepackage{hyperref}
\usepackage[dvipsnames]{xcolor}
\usepackage{longfbox}
\usepackage{multirow}
\usepackage{mdframed}
\usepackage{lipsum}
\usepackage{graphicx}
\usepackage{setspace}

\title{KRAG Framework for Enhancing LLMs in the Legal Domain}

\author{%
Nguyen Ha Thanh\textsuperscript{1,2}\and
Ken Satoh\textsuperscript{1} \\
\affiliations
\textsuperscript{1}Center for Juris-Informatics, ROIS-DS, Tokyo, Japan\\
\textsuperscript{2}Research and Development Center for Large Language Models, NII, Tokyo, Japan\\
\emails
\{nguyenhathanh, ksatoh\}@nii.ac.jp
}

\begin{document}

\maketitle

\begin{abstract}
  This paper introduces Knowledge Representation Augmented Generation (KRAG), a novel framework designed to enhance the capabilities of Large Language Models (LLMs) within domain-specific applications. KRAG points to the strategic inclusion of critical knowledge entities and relationships that are typically absent in standard data sets and which LLMs do not inherently learn. In the context of legal applications, we present Soft PROLEG, an implementation model under KRAG, which uses inference graphs to aid LLMs in delivering structured legal reasoning, argumentation, and explanations tailored to user inquiries. The integration of KRAG, either as a standalone framework or in tandem with retrieval augmented generation (RAG), markedly improves the ability of language models to navigate and solve the intricate challenges posed by legal texts and terminologies. This paper details KRAG's methodology, its implementation through Soft PROLEG, and potential broader applications, underscoring its significant role in advancing natural language understanding and processing in specialized knowledge domains.

\end{abstract}

\section{Introduction}
Large language models (LLMs) have seen rapid advancements in recent years \cite{radford2019language,brown2020language,ouyang2022training,manyika2023overview,team2023gemini,jiang2023mistral,achiam2023gpt} demonstrating impressive zero-shot learning and performance across various natural language processing tasks. Prominent among the capabilities of LLMs is their capacity to extract information from their pre-trained parameters. However, these models still struggle with certain challenges, such as hallucination of information and limited precision in manipulating knowledge \cite{lewis2020retrieval}.

\begin{figure}
    \centering
    \includegraphics[width=0.5\textwidth]{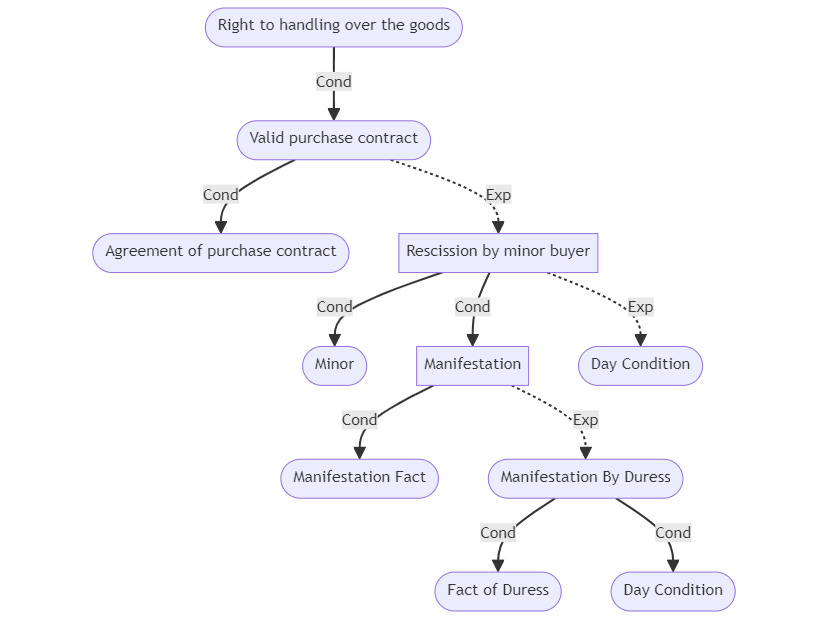} 
    \caption{A graph explaining legal situations created by SoftPROLEG, highlighting the importance of guiding the model in making decisions using a well-structured knowledge representation.}
    \label{fig:softproleg_example}
\end{figure}

Retrieval-augmented generation (RAG) is a technique that aims to mitigate some of these issues by combining pre-trained parametric models with non-parametric memory \cite{lewis2020retrieval}. RAG techniques have been shown to support LLMs to generate more specific, diverse, and factual language than traditional seq2seq \cite{sutskever2014sequence} baselines. Nevertheless, certain aspects of knowledge representation remain underdeveloped, particularly in domains requiring specialized knowledge.

While RAG addresses some limitations, prompting techniques offer another avenue to enhance LLM reasoning capabilities.
Prompting techniques, such as chain of thought (CoT) prompting, have been proposed to enhance LLM reasoning capabilities \cite{wei2022chain}. By providing a few exemplar reasoning steps, these methods can guide the model towards more accurate and coherent responses. This approach has shown promising results in tasks involving arithmetic, commonsense, and symbolic reasoning. However, the complexity of real-world domains like law (see Figure \ref{fig:softproleg_example} as an example), where multiple principles may simultaneously govern social relationships, necessitates a more fine-grained knowledge representation approach for LLMs \cite{otero2023orla,rago2023interactive}, along with an educated conflict resolution mechanism to ensure consistent judgments.

In high-stakes domains, explainable and dependable models are essential for making decisions based on accurate and understandable information \cite{leofante2023counterfactual}. In legal settings, even large language models equipped with RAG and advanced prompting strategies, such as enhancements, still produce inconsistent rulings due to inadequacies in their knowledge representation (see Section \ref{sec:related_work}). Judges and lawyers, through professional training and experience, develop a well-structured knowledge base and understanding of legal nuances. To achieve accurate and reliable decision-making in complex law and policy domains, it is crucial to devise a method that focuses on improving knowledge representation. 

This paper introduces the Knowledge Representation Augmented Generation (KRAG) framework, specifically tailored to boost the performance of LLMs in complex legal contexts through Soft PROLEG, our implementation model. By integrating structured knowledge representations like inference graphs, KRAG enhances LLMs' comprehension of legal intricacies and improves their output precision. Throughout this study, we articulate the KRAG methodology and its synergistic integration with existing techniques like RAG and CoT and demonstrate its efficacy via empirical analyses. We also discuss broader applications and future prospects of KRAG in specialized domains.


\section{Related Work}
\label{sec:related_work}

One line of research focuses on designing benchmarks and evaluating the legal knowledge of LLMs. For example, \cite{fei2023lawbench} proposes a comprehensive benchmark, LawBench, to evaluate LLMs' legal capabilities across multiple cognitive levels, including legal knowledge memorization, understanding, and application. Additionally, \cite{nguyen2023negation} introduces a benchmark specifically for assessing state-of-the-art models in the legal domain on the task of negation detection. Although these studies evaluate a variety of LLMs' performances on diverse legal tasks, they do not explicitly dive into the mechanisms through which LLMs can better leverage structured knowledge, like knowledge graphs. Exploring such mechanisms could be a promising research direction.


Another branch of research is dedicated to adapting LLMs for specific legal tasks. \cite{huang2023lawyer} presents a framework for adapting LLMs to the legal domain and builds Lawyer LLaMA, a legal domain LLM. While this study demonstrates the feasibility of injecting domain knowledge during the training stage and using supervised fine-tuning tasks to teach professional skills, it does not focus on how to incorporate knowledge graphs or other structured knowledge sources to enhance LLMs' legal reasoning abilities further.


There is a growing interest in unifying LLMs and knowledge graphs (KGs) to benefit from each other's strengths. For instance, \cite{pan2024unifying} discusses a forward-looking roadmap for this unification, including three general frameworks: KG-enhanced LLMs, LLM-augmented KGs, and synergized LLMs + KGs. However, this paper serves primarily as a general roadmap and overview, leaving considerable room for research into specific techniques and methodologies for integrating LLMs and KGs.


Several works address the integration of legal knowledge in representation learning. \cite{xiao2023legal} focuses on legal knowledge-intensive AI approaches, providing a summary of the existing methods for knowledge representation, acquisition, and application. Similarly, \cite{suchanek2023knowledge} emphasizes the complementarity of LLMs and structured data repositories like KGs. These works offer valuable perspectives on leveraging legal knowledge for AI-based legal task-solving. However, they do not provide specific techniques for enhancing LLMs' legal reasoning with structured knowledge, leaving this area open to exploration.


Recent studies, such as \cite{chang2023survey}, highlight the importance of evaluating LLMs across various tasks and ethical dimensions. Additionally, \cite{alexopoulos2023natural} discusses the potential of NLP-powered legal applications in India, while \cite{belle2023knowledge} looks into the challenges and opportunities in ethical AI, for knowledge representation and acquisition. While these research efforts contribute to the discussion on the evaluation and application of LLMs from various ethical and task-specific perspectives, they do not specifically address their application or enhancement in the legal domain.

\section{Preliminary}

\subsection{Presupposed Ultimate Fact Theory}
The Presupposed Ultimate Fact
Theory (JUF theory) \cite{ito2008lecture} is designed to help allocate the burden of proof in civil law cases, particularly in situations where complex legal claims, counterclaims, and ultimate facts must be considered based on available evidence. Here, we use a hypothetical scenario involving a goods sale contract dispute to illustrate the JUF theory. We designate the characters and events involved in this scenario with specific symbols for clarity.

In this dispute, a buyer \(B\) claims that they entered into a sales contract \(C\) with the seller \(S\), who failed to deliver the goods \(G\) as per the agreement. The buyer \(B\) demands a refund \(R\) along with compensation \(D\) for additional costs incurred due to the seller's \(S\) breach of contract. In defense, the seller \(S\) counterclaims that they shipped the goods \(G\) on time and any delivery delays \(L\) were beyond their control (e.g., due to logistical issues or force majeure \(F\)). Moreover, the seller \(S\) contends that the buyer \(B\) had approved additional delivery time \(T\) during their initial communication.

To resolve such cases, it is critical to determine which party is responsible for providing evidence to support their respective claims and counterclaims. The JUF theory offers guidance on allocating \textit{the burden of proof} for each ultimate fact. In this example, both the buyer \(B\) and the seller \(S\) bear the burden of proof for their respective claims \((C_{B}, R_{B}, D_{B})\) and counterclaims \((G_{S}, L_{S}, F_{S}, T_{S})\). If either party fails to prove their claim, it is considered false, leading the court to make a decision accordingly.

In practice, situations may arise where the validity of a contract \(C\) is questioned due to factors such as one party being underage, the presence of coercion, or a breach of contract conditions. In such cases, the allocation of the burden of proof plays a crucial role in determining the outcome of a legal dispute. The JUF theory provides a robust framework to assess the veracity of each claim by assigning the responsibility to produce evidence and establish the truthfulness of ultimate facts.

\subsection{PROLEG: A Logic-Based Approach}
Addressing the challenges of automating the JUF theory, PROLEG \cite{satoh2010proleg} offers a logic-based solution that employs a formal rule representation and fact-handling mechanism. PROLEG modifies the conventional logic programming method to provide a user-friendly and intuitive rule format while incorporating the JUF theory's openness concept, which distinguishes between normal and exceptional ultimate facts.

To make the rules more comprehensible to users, PROLEG separates the positive condition part and the negation-as-failure part in the rule representation. It introduces an intermediate concept, which aggregates meaningful sets of ultimate facts or other intermediate concepts, enhancing readability and simplifying the representation of complex legal scenarios. Additionally, PROLEG introduces new predicates, such as ``exception,'' to represent exceptional situations that may alter the conclusion.

When handling facts, PROLEG employs four types of predicates: ``allege,'' ``provide evidence,'' ``plausible,'' and ``admission.'' These predicates correspond to the actions performed by each party during a legal argument at the court as well as judge's decision-making processes. They collectively model the burden of production, providing evidence and admission of facts in civil litigation. By incorporating these predicates into its reasoning process, PROLEG effectively guides the allocation of the burden of proof and evaluates the truth or falsity of ultimate facts as required in legal disputes.

As a result, PROLEG's formal representation and fact-handling mechanism enables it to capture the intricacies, subtleties, and complexities of legal reasoning and argumentation. This logic-based approach empowers the PROLEG engine to simulate the JUF theory's inference accurately, providing a valuable tool for automating legal reasoning processes. However, this logic-dependent method also presents challenges in handling and interpreting natural language forms.

\subsection{Translation-based PROLEG Approach}

Recent studies \cite{nguyen2022multi,zin2023improving} have sought to bridge the gap between the complexities of natural language and the structured demands of legal reasoning frameworks like PROLEG through the development of translation methodologies. These methodologies focus on converting natural language descriptions of legal scenarios into the fact descriptions required by the PROLEG system.  Allowing facts to be described in natural language, which is then automatically translated into the system's required format, significantly lowers the barrier for non-experts to engage with legal reasoning technologies. This could lead to broader adoption and utilization of legal reasoning systems in everyday legal inquiries and disputes, empowering individuals with tools previously accessible only to legal professionals with specialized training.

However, several challenges and limitations accompany this translation approach. Firstly, despite facilitating the input of facts in natural language, the underlying rule structure of PROLEG remains unchanged and must still be manually crafted by experts. This rigidity means that while the input process becomes easier for lay users, the development and modification of the legal reasoning framework itself remains highly technical and inaccessible to most users. Secondly, translating natural language into fact descriptions does not guarantee that the resulting facts will align perfectly with the existing rules within the system. Discrepancies between the nuanced meanings inherent in natural language texts and the strict logic required by rule-based systems can lead to misapplications or incorrect interpretations of rules, potentially compromising the accuracy of legal reasoning outcomes. 

\subsection{Legal Debugging}

Legal debugging \cite{fungwacharakorn2022debugging} refers to the process of identifying, analyzing, and correcting unexpected consequences that arise from the application of legal reasoning systems. In the context of PROLEG, this involves the examination of instances where the system's interpretation of law, based on its codified factors, fails to consider new or exceptional circumstances, resulting in erroneous or unjust outcomes.

The dynamic nature of legal cases often brings new and unforeseen factors into consideration, presenting a significant challenge for legal reasoning systems like PROLEG. Given that PROLEG, like many legal frameworks, operates on the basis of predefined legal rules and codified factors, it is primarily equipped to apply the law in its plain meaning. The reliance on expressly defined conditions means that the system may not fully account for novel factors, potentially leading to outcomes that don't align with the evolving landscape of legal reasoning and justice. 

Given the static nature of codified rules within systems like PROLEG, the incorporation of new legal factors or exceptional cases often requires significant manual intervention. This manual process entails identifying the oversight, understanding the implications of the new factor, and revising the legal rules within the system to encompass the exception—a process that is both time-intensive and demands a high degree of legal expertise.

\subsection{Advanced Prompting Techniques \& KRAG}

In the context of legal applications, leveraging advanced prompting techniques can enhance the performance and effectiveness of large language models (LLMs) in understanding, reasoning, and generating solutions for legal problems. 

\begin{enumerate}
  \item \textbf{IO Prompting (Zero-Shot):} In legal applications, IO prompting guides the LLM to analyze the legal problem $p$ and produce an appropriate solution $s = f_\theta(p)$ based on the model's pre-trained knowledge and reasoning capabilities.
    
  \item \textbf{One-Shot:} By incorporating a single example $x$ during inference, the one-shot methodology allows the LLM to draw parallels between the given example and the legal problem under consideration, enhancing its problem-solving capacity as $s = f_\theta(p, x)$.
  
  \item \textbf{Generated Knowledge Prompting:} This technique involves the LLM generating relevant legal knowledge $k = f_\theta(p)$ in response to a given problem, which is then used as contextual information to form a well-informed solution: $s = f_\theta(p, k)$.
  
  \item \textbf{Chain-of-Thought (CoT) Prompting:} CoT decomposes legal reasoning into a multi-step process, with each sub-problem being addressed sequentially—$z_i = f_\theta(p; \{z_j | j < i\})$—ultimately leading to a comprehensive solution to the overarching legal problem.
  
  \item \textbf{Self-Consistency CoT:} In legal scenarios, self-consistency CoT aggregates predictions and selects a solution based on the consensus of multiple outputs generated from different prompts of the same task: $s = maj\{f_\theta(p_1), f_\theta(p_2),\dots\}$.
    
  \item \textbf{Tree-of-Thought (ToT) Prompting:} ToT prompting enables the LLM to evaluate and select the best solution candidate at each step of the legal reasoning process, with the model proposing a set of sub-solution candidates $\{z^n_{i+1}|n = 1 \dots N\} = f_\theta(p; \{z_j | j \leq i\})$ before assessing the most suitable course of action.
\end{enumerate}

These advanced prompting techniques demonstrate that large language models (LLMs) can achieve enhanced performance when given appropriate prompts. However, without well-implemented guidelines, these techniques can fall short in several critical areas, including the incorporation of intricate contextual nuances, the detailed decomposition of complex problem structures, and the ability to draw contextually and accurately from a vast repository of domain-specific knowledge.

To address these limitations, a more comprehensive approach is needed, which not only leverages advanced prompting techniques but also integrates robust knowledge representation and retrieval mechanisms. This leads us to the concept of Knowledge Representation Augmented Generation (KRAG).

KRAG represents an overarching framework that goes beyond traditional prompting techniques by combining knowledge representation and retrieval with generation processes. This dual capability allows the system to draw on precise, structured knowledge while engaging in the creative generation of solutions tailored to specific contexts. Here's how KRAG integrates and enhances these elements:

\[ \text{KRAG}(p) = g\Big(f_\text{retrieval}(p, \mathcal{K}), f_\text{structure}(p, \mathcal{G})\Big) \]

\noindent where:
\begin{itemize}
    \item \( p \) is the problem.
    \item \( \mathcal{K} \) and \( \mathcal{G} \) represent the knowledge base of past experiences and domain-specific information and structured guidelines.
    \item \( f_\text{retrieval}(p, \mathcal{K}) \) is the function responsible for retrieving relevant knowledge and past examples.
    \item \( f_\text{structure}(p, \mathcal{G}) \) structures the problem into manageable components to guide the reasoning process.
    \item \( g \) is the generation function that integrates retrieval and representation outputs to produce the solution.
\end{itemize}

The legal domain is an ideal testbed to experiment with this concept, and we introduce Soft PROLEG as a specific implementation of KRAG in the following section.

\section{Soft PROLEG}

\subsection{Derivational Analogy in Soft PROLEG}

\begin{figure} 
    \centering
    \includegraphics[width=0.45\textwidth]{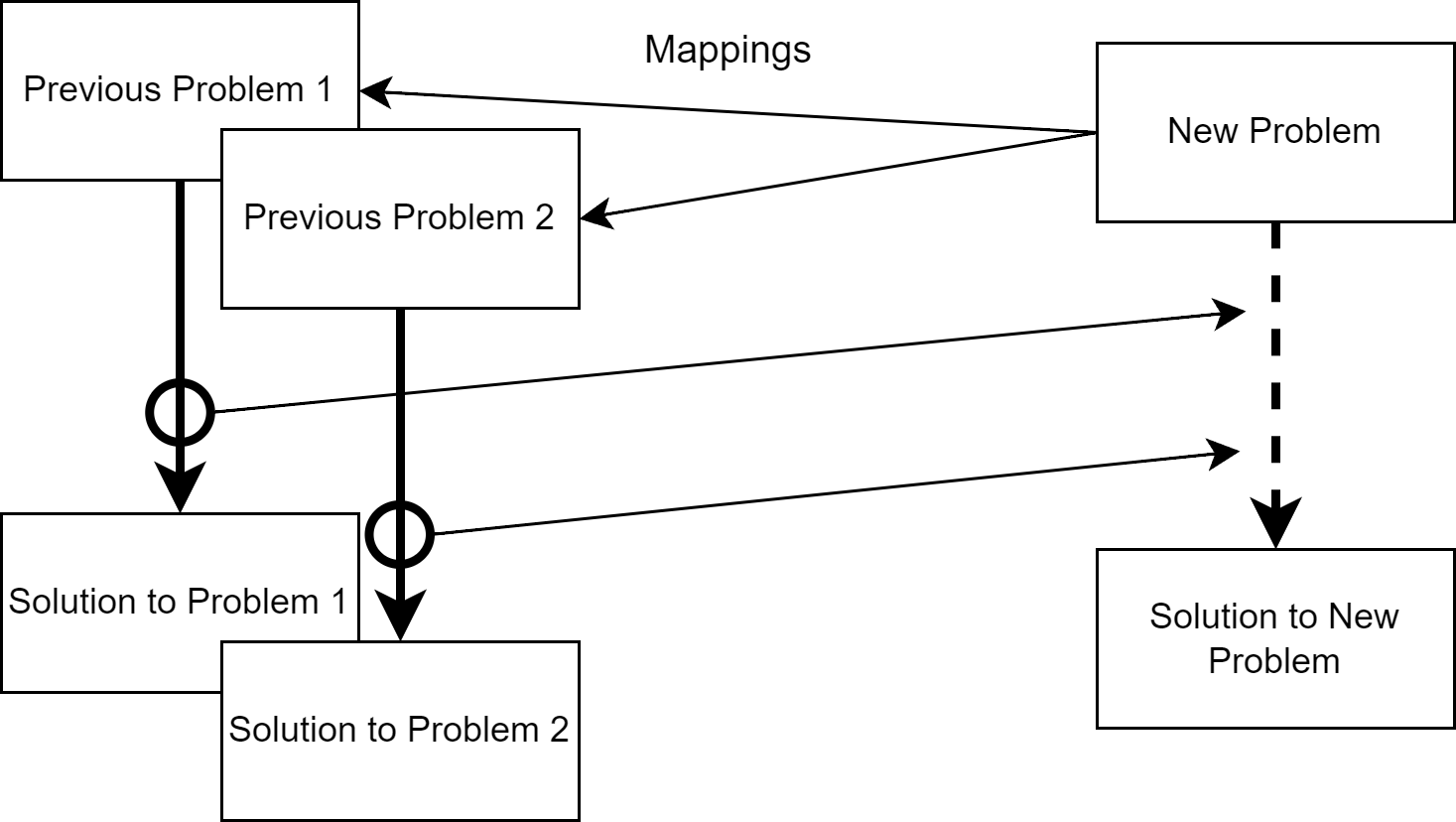} 
    \caption{Derivational Analogy.} 
    \label{fig:derivational_analogy} 
\end{figure}

The concept of derivational analogy (Figure \ref{fig:derivational_analogy}) \cite{carbonell1985derivational} serves as a theoretical foundation for the development of Soft PROLEG, which aims to address complex problem-solving situations, such as legal structures. Unlike transformational analogy, which involves altering and adapting past solutions to fit new problems, derivational analogy focuses on reconstructing problem-solving strategies based on previous experiences. This approach is more appropriate for Soft PROLEG, as it requires a deeper level of understanding and flexibility when handling intricate legal scenarios.

In the context of KRAG, and specifically within civil law, \( \mathcal{K} \) corresponds to the relevant legal texts and statutes, while \( \mathcal{G} \) represents the method of applying these legal texts to specific situations. This application is guided by the interpretation and disposition of legal factors within both the question at hand and the applicable laws. By employing KRAG, Soft PROLEG ensures that complex legal scenarios are systematically deconstructed and addressed using a robust combination of knowledge retrieval and structured reasoning.

The approach focuses on dividing legal conditions into smaller subconditions and accounting for relevant exceptions, allowing the LLM to learn and comprehend legal texts with greater precision. The methodology can be formalized as follows:

\[ A = SoftPROLEG (C_1, C_2, ... , C_n; E) \]

Where \( A \) represents the main legal condition, \(C_1, C_2, ... , C_n\) are the subconditions that need to be satisfied for \(A\) to hold true, and \(E\) is the exception that negates the fulfillment of \(A\) if satisfied.

By structuring problems into main conditions and subconditions, Soft PROLEG systematically simplifies complex legal issues, aligning closely with the principles of the JUF theory, which emphasizes the allocation of the burden of proof in legal disputes. This structured approach enhances the ability to identify similarities across different legal scenarios. For example, LLMs can leverage their extensive knowledge base to detect patterns and commonalities in the formulation of \( C_1, C_2, ... , C_n \) across varied legal databases and texts. Such recognition aids in applying relevant historical knowledge to new, comparable cases, closely reflecting the JUF theory's guideline on how evidence should be assessed and utilized based on past decisions.

Facing a new legal scenario, LLMs utilize this training to identify which subconditions \( C_1, C_2, ... , C_n \) were pertinent in akin previous cases. This capability to retrieve and apply historical knowledge is vital for efficient problem-solving within legal contexts, where resolutions often depend on precedents and the similarities to earlier cases, echoing the JUF theory's focus on the evidentiary basis for legal decisions.

The structure \( A = SoftPROLEG (C_1, C_2, ... , C_n; E) \) enables LLMs to operationalize knowledge transfer by extracting applicable information from preceding cases and integrating it into the specific subconditions of a new case. This model facilitates a methodical approach to compare, adapt, and amalgamate previous legal solutions to tackle new challenges, considering all pertinent aspects and exceptions, which is a core component of the JUF theory's application in practical legal contexts.

Utilizing structured data such as \(C_1, C_2, ..., C_n; E\), LLMs can proficiently navigate their extensive databases to uncover prior instances where similar conditions and exceptions were managed. Advanced algorithms then evaluate these occurrences for their relevance to the current case, optimizing the search for and retrieval of the most pertinent precedents, thus enabling a practical application of the JUF theory's principles in automating legal reasoning and decision-making processes.



\subsection{Graph Structure}

The graph representation consists of the following components:

\begin{enumerate}
    \item Nodes: Each node corresponds to a legal condition (\(C\)) or an exception (\(E\)). Nodes can be further classified into two types:
    \begin{itemize}
        \item Condition Nodes (\(C\)): Represent subconditions that need to be satisfied for a given main legal condition.
        \item Exception Nodes (\(E\)): Represent exceptions that invalidate the fulfillment of the main legal condition if satisfied.
    \end{itemize}
    
    \item Edges: Edges in the graph represent relationships between connected nodes. They signify the dependencies among subconditions and exceptions. An edge between nodes can be denoted as (\(C_i, C_j\)) or (\(C_i, E_j\)), indicating that the fulfillment of condition \(C_i\) is dependent on the satisfaction of condition \(C_j\) or exception \(E_j\).
\end{enumerate}

\subsection{Knowledge Set Construction}

The graph can be constructed using diverse, representative, and labeled examples of legal texts that cover various conditions, subconditions, and exceptions. Each example may represent a unique scenario or legal context. To build the graph, the following steps can be performed:

\begin{enumerate}
    \item Identify and represent the main legal condition (\(A\)) using the Soft PROLEG methodology.
    \item Map each subcondition (\(C_i\)) and exception (\(E_j\)) to nodes in the graph.
    \item Determine relationships between the identified nodes by examining the dependencies among the subconditions and exceptions in the legal text and connect the dependent nodes with edges.
    \item Add the graph to the knowledge base, which contains graphs for various other legal contexts. If any of the nodes or edges overlap with existing graphs, unify those graphs accordingly.
\end{enumerate}

In practice, with this method, we can create a knowledge base in the form of a forest or tree structure. By traversing this structure, the LLM can easily provide accurate decisions as well as visualize and explain the decisions to users in an intuitive and comprehensible manner. Moreover, if a cycle is generated during the graph creation process, it may indicate the presence of an ambiguous or contradictory rule. This, in turn, supports the legal debugging process mentioned in the previous section.





\section{PoC Implementation and Evaluation}
This section delineates the architecture and operational mechanics of the Proof of Concept (PoC) system developed for Soft PROLEG 1.0.

\subsection{System Architecture}

\begin{figure*} 
    \centering
    \includegraphics[width=0.9\textwidth]{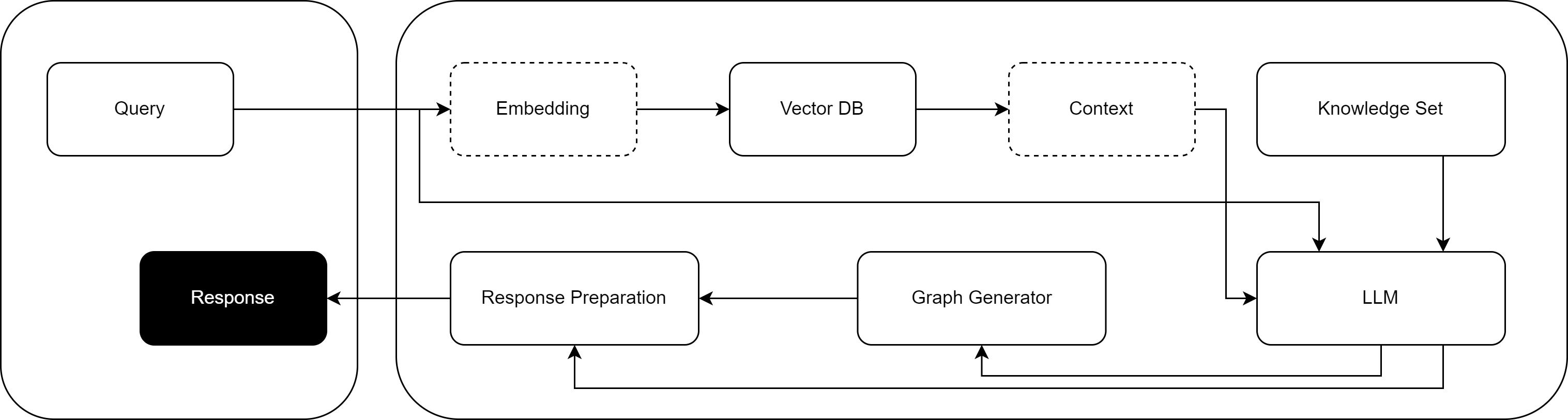} 
    \caption{General architecture of Soft PROLEG 1.0} 
    \label{fig:architecture} 
\end{figure*}

Figure \ref{fig:architecture} presents a comprehensive flowchart that illustrates the architecture of our KRAG PoC system (i.e. SoftPROLEG 1.0). At the outset, this architecture is designed to streamline the process from the initial user query to the final delivery of a response, leveraging techniques in data representation, search, and natural language processing. 

\textbf{Query Submission}: The process commences when the user submits a query. This query outlines the user's information needs or the problem they aim to resolve using the system.

\textbf{Embedding Conversion}: Once the user's input is received, the system transforms the query into an embedding. An embedding is a numerical representation of the query, typically expressed as a vector in a multi-dimensional space, and it's designed to facilitate the retrieval process.

\textbf{Vector Database Search}: After the query has been converted into an embedding, it is used to perform a search within a vector database (DB). The goal of this step is to identify any legal regulations related to the query.

\textbf{Context Retrieval}: The system retrieves the relevant legal articles from the vector DB. The retrieval algorithm can range from similarity comparisons to bi-encoder or cross-encoder retrieval methods.

\textbf{Knowledge Set}: At this point, the system has the query and context, as with any RAG application. However, the distinctive feature of the KRAG model is the knowledge set. This set is a repository of pattern graphs that are used to analyze and apply the query and context.

\textbf{Large Language Model}: The LLM undertakes the role of understanding the query and context, and identifying the most similar graph for the current scenario. The results are then forwarded to the graph generator and response preparation modules.

\textbf{Graph Generation}: Based on the information received from the LLM, this module generates a graph to explain the LLM's decision. This step enhances the transparency of the reasoning process and aids troubleshooting during the legal debugging phase.

\textbf{Response Preparation}: This module's responsibility is to prepare the user's response based on the LLM's decision, the generated graph, and related content (e.g., disclaimer, additional annotation).

\textbf{Response Delivery}: In the final step, the system presents its response to the user. This response is the culmination of the system's comprehensive processing and is designed to provide a well-informed and clearly articulated answer that directly addresses the initial query.

\subsection{Knowledge Set Construction}

The construction of the Knowledge Set is paramount to the effectiveness of the KRAG system in SoftPROLEG 1.0, as it serves as the core repository of legally educated precedents utilized by the Large Language Model (LLM) to apply knowledge precisely and appropriately to each query. 

\subsubsection{Semi-Automated Process}

The process of constructing the Knowledge Set for the KRAG system is both intricate and dynamic, involving a semi-automated method that leverages the sophisticated capabilities of LLMs while ensuring reliability and accuracy through human verification:

\begin{enumerate}
    \item \textbf{Initial Data Generation by LLM:} Utilizing the robust analytical capabilities of LLMs, an initial set of data points is generated. These data points are derived from scenarios and questions resembling those found in the Japanese Bar Exam \cite{goebel2024overview}.

    \item \textbf{Human Expert Review and Verification:} To maintain the accuracy and applicability of the Knowledge Set, each entry generated by the LLM is reviewed and verified by legal experts. This step is crucial to ensure that the information adheres strictly to legal standards and practical applicability.

    \item \textbf{Augmentation and Enrichment:} Post-verification, the data is further enriched with additional information, annotations, or modifications as suggested by human experts. This ensures a richer and more detailed set of legal precedents for the system to utilize.
\end{enumerate}

\subsubsection{Structural and Representational Approach}

Each entry within the Knowledge Set contains several components that structure and contextualize the legal information, facilitating effective knowledge retrieval and application. Each knowledge point, \( K_i \), is defined as:

\[ K_i = (Q, R, G) \]

\noindent Where:
\begin{itemize}
    \item \( Q \) represents the \textbf{query}, mirroring potential inquiries that could be raised by users, formulated based on actual questions from the Japanese Bar Exam to ensure realism and relevancy.
    \item \( R \) includes \textbf{related articles} or legal documents pertinent to \( Q \). These documents are critical as they provide the foundational laws, previous case decisions, and annotations essential for understanding and addressing the query.
    \item \( G \) denotes the \textbf{graph structure}, which encompasses nodes and edges derived and decomposed from the query \( Q \) and related articles \( R \). This graph presents the main legal constructions, subconditions, and any exceptions.
\end{itemize}

\subsubsection{Implementation and Utilization}

\begin{figure} 
    \centering
    \includegraphics[width=0.45\textwidth]{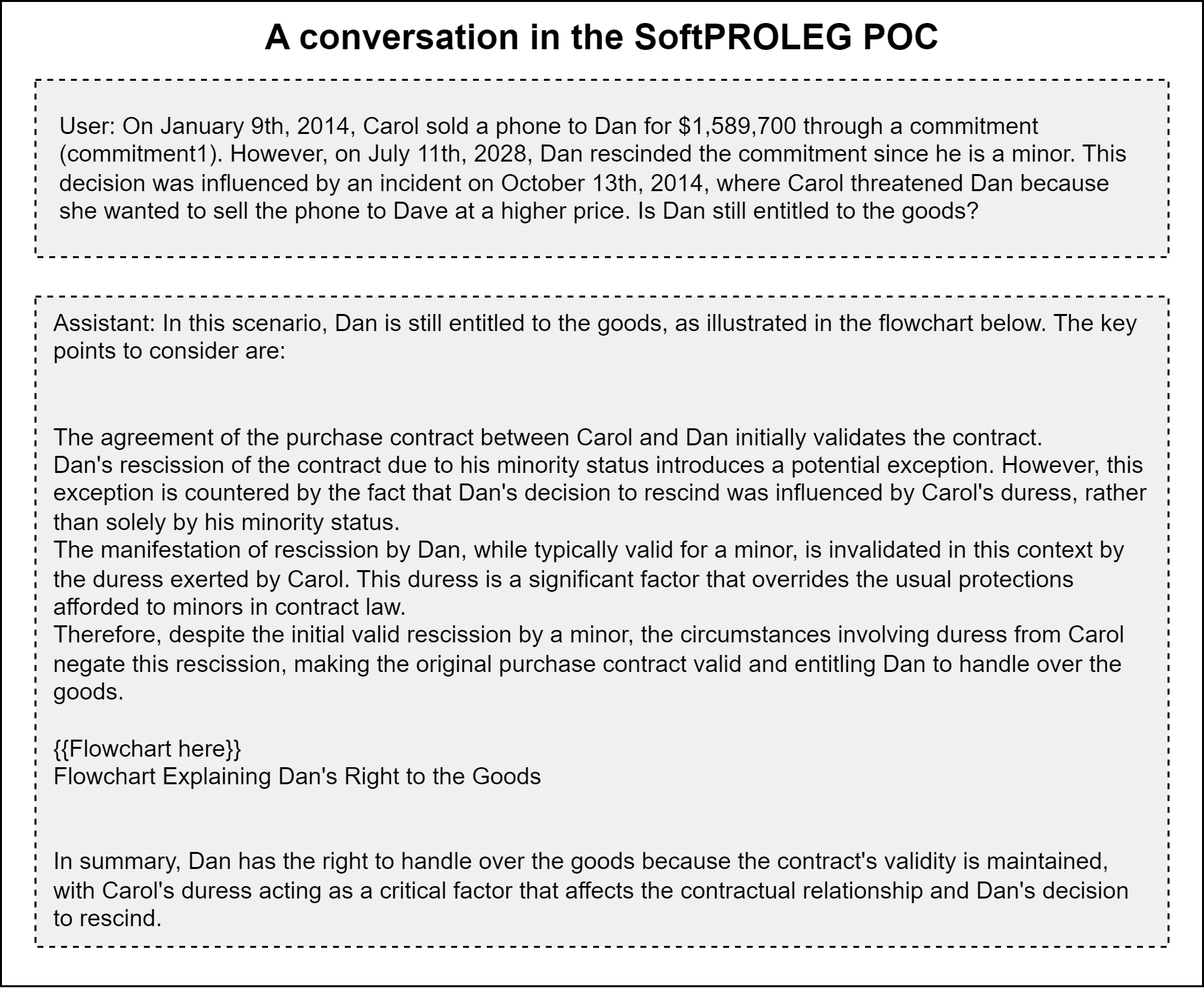} 
    \caption{An example of a conversation with the Soft PROLEG System, as compared to a vanilla LLM, shows that the response is clear, logical, and legally sound, accompanied by an illustration (the flowchart shown in Figure \ref{fig:softproleg_example}).} 
    \label{fig:example} 
\end{figure}

The initial Proof of Concept (PoC) version 1.0 of the system integrated 1,287 samples into the Knowledge Set, each corresponding to specific instances of legal queries akin to those found in the Japanese Bar Exam. This comprehensive database serves as a fundamental component in the KRAG system, offering a wide array of scenarios for LLM to apply and adapt its learned legal knowledge effectively.
An example of how the system processes such queries can be seen in Figure \ref{fig:example}.

In the current implementation, the retrieval and application of knowledge from the Knowledge Set leverage both lexical and vector-based search techniques. This approach ensures efficient matching of user queries with relevant legal precedents. However, looking ahead to future iterations of the KRAG system, there are plans to incorporate graph search methodologies.
This method would involve constructing a graph where each node represents a condition \( C_i \) or a legal precedent found in the Knowledge Set, providing a more interconnected and dynamic method of querying the database. This enhancement will likely contribute significantly to the precision and adaptability of the KRAG system in dealing with complex legal inquiries.

\subsection{Evaluation}

\subsubsection{Experimental Setup}

The objective of the evaluation is to demonstrate the utility of the proposed solution through two attributes: performance and stability. This involves answering two key questions:
\begin{enumerate}
    \item Does the KRAG approach allow the SoftPROLEG system to outperform non-KRAG systems?
    \item Does the KRAG approach enhance the stability of the SoftPROLEG system compared to non-KRAG systems?
\end{enumerate}

To achieve this, we conducted experiments using the English version of the Japanese Bar Exam over five years, from Heisei 29 (2017) to Reiwa 03 (2021).The backbone model for SoftPROLEG utilizes both GPT-3.5 and GPT-4, and we compared the performance and stability characteristics of SoftPROLEG to the original versions of both models. The choice of GPT-3.5 as a backbone helps highlight the role of the KRAG system in enhancing the model's reasoning capabilities, given the original GPT-3.5's limited performance in reasoning \cite{nguyen2023negation,sanyal2024machines,agrawal2024exploring}.

Each system/model was tasked with answering a set of questions twice, allowing us to extract data on their performance and stability. The measure for performance is the average accuracy across the two instances, and the measure of stability is the percentage of questions for which the model provided the same answer in both rounds. These metrics are defined as follows:

\begin{equation}
    \text{Performance} = \frac{\text{Number of Correct Answers in both trials}}{\text{Total Number of Questions}}
\end{equation}

\begin{equation}
    \text{Stability} = \frac{\text{Number of Consistent Answers in both trials}}{\text{Total Number of Questions}}
\end{equation}

\begin{figure} 
    \centering
    \includegraphics[width=0.45\textwidth]{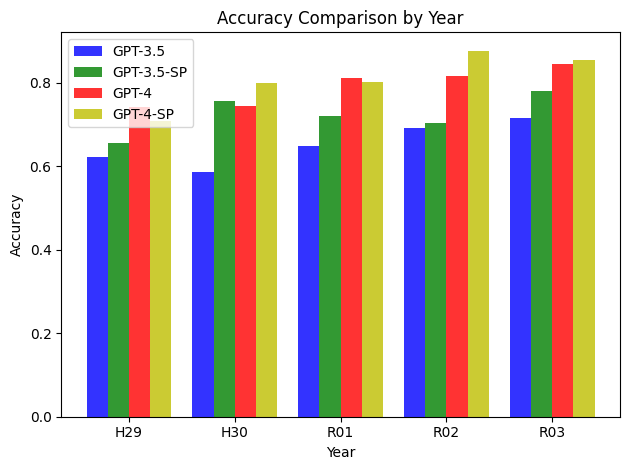} 
    \caption{Comparison of accuracy across different models: GPT-3.5, GPT-3.5-SP (SoftPROLEG with a GPT-3.5 backbone), GPT-4, and GPT-4-SP (SoftPROLEG with a GPT-4 backbone). This bar chart illustrates the differing levels of accuracy each model achieved on the English version of the Japanese Bar Exam spanning from Heisei 29 (2017) to Reiwa 03 (2021). The results highlight the impact of the KRAG system on enhancing the reasoning abilities of the underlying GPT-3.5 and GPT-4 models in the SoftPROLEG implementations.} 
    \label{fig:accuracy} 
\end{figure}




\begin{figure*} 
    \centering
    \includegraphics[width=0.9\textwidth]{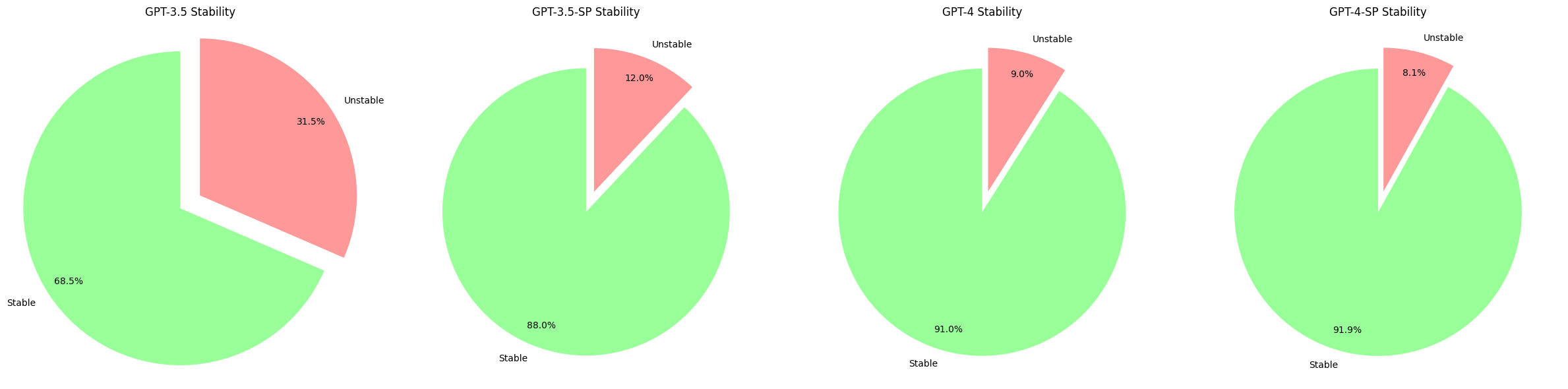} 
    \caption{Stability analysis of GPT-3.5, GPT-3.5-SP (SoftPROLEG with a GPT-3.5 backbone), GPT-4, and GPT-4-SP (SoftPROLEG with a GPT-4 backbone) through pie charts. From left to right, the charts display the consistency of responses provided by each model across two trials of the same set of questions from the English version of the Japanese Bar Exam. These visual representations underscore how the integration of the KRAG system in GPT-3.5-SP and GPT-4-SP enhances response consistency compared to both the baseline GPT-3.5 and the more advanced GPT-4 model.} 
    \label{fig:stability} 
\end{figure*}


\subsubsection{Experimental Results}

The experimental results are summarized in Figure \ref{fig:accuracy} and Figure \ref{fig:stability}, displaying the accuracy and stability of each model respectively. The accuracies achieved by GPT-3.5, GPT-3.5-SP, GPT-4, and GPT-4-SP are as follows:

\begin{itemize}

    \item GPT-3.5 recorded accuracies of 0.6207, 0.5857, 0.6486, 0.6915, and 0.7156 over the five-year span of the bar exam.

    \item GPT-3.5-SP showed improved accuracies with values of 0.6552, 0.7571, 0.7207, 0.7037, and 0.7798, indicating the efficacy of the KRAG system in enhancing reasoning capabilities.

    \item GPT-4 demonstrated high accuracies, ranging from 0.7414 to 0.8440, reflecting its advanced NLP capabilities.

    \item GPT-4-SP exhibited strong performance with accuracies of 0.7069, 0.8000, 0.8018, 0.8765, and 0.8532, further showcasing the benefits of the KRAG system.

\end{itemize}

Stability percentages for each model were as follows:

\begin{itemize}

    \item GPT-3.5 showed a stability of 68.5\%.

    \item GPT-3.5-SP exhibited a significant increase in stability, reaching 88.5\%.

    \item GPT-4 maintained high stability at 91\%.

    \item GPT-4-SP achieved the highest stability at 91.9\%.

\end{itemize}

The data reveals that both GPT-3.5-SP and GPT-4-SP outperform the original models in terms of accuracy and stability. Although GPT-4 shows robust performance in accuracy overall (consistent with the results reported by \cite{katz2023gpt}), the KRAG system's contribution to additional gains in both stability and accuracy is remarkable. A significant gap is particularly evident in Reiwa 02 (2020), where GPT-4-SP considerably outperforms GPT-4, indicating that KRAG enhances logical reasoning and consistency. This underscores the KRAG system's potential to further improve the performance of even the most advanced models, making it a valuable asset for applications requiring high reliability and precision in responses.

The superior stability noted in GPT-3.5-SP and GPT-4-SP suggests that the KRAG approach not only improves immediate reasoning based on single queries but also ensures consistency across repeated inquiries, a critical factor in legal and professional settings.

Moreover, the prompts within the SoftPROLEG system require the model to generate explanatory text and corresponding graphical structures, as illustrated in Figures \ref{fig:softproleg_example} and \ref{fig:example}. This feature showcases the superior approach of the system in terms of performance, stability, and practical applicability.

\section{Discussion}

\subsection{Implications of Soft PROLEG for Legal Reasoning and LLMs}

The introduction of Soft PROLEG under the KRAG framework represents a significant advancement in the application of Large Language Models (LLMs) in the domain of legal reasoning. This system not only enhances the precision of LLMs in understanding and generating legal content but also provides a structured mechanism for incorporating complex legal rules and context into the decision-making process. By integrating knowledge representation with augmented generation, Soft PROLEG allows for more nuanced and legally coherent outputs, which are crucial for legal applications where the stakes of incorrect information are high.

Furthermore, the use of linked graphs to break down legal queries into components such as conditions, sub-conditions, and exceptions introduces a level of interpretability and transparency previously challenging to achieve. This advancement leads to better trust and reliability in automated legal advice, potentially transforming how professionals and the public interact with legal services.

\subsection{Limitations and Areas for Improvement}

While Soft PROLEG showcases considerable improvements in the application of LLMs to legal reasoning, several limitations warrant attention. One primary concern is the computational complexity involved in handling large knowledge graphs and performing real-time inference. This concern stems from the fact that LLMs are inherently resource-intensive. However, with the rapid advancements in LLM technology, this issue is likely to be resolved soon. Therefore, it is reasonable to prioritize the accuracy and utility of the system over its speed within a balanced consideration.

Additionally, the current implementation relies heavily on a semi-automated process for knowledge set construction, involving substantial human effort in verifying and enriching data points. This process, while crucial for accuracy, limits the scalability of the system. Automating more aspects of this process without compromising the quality of the legal knowledge base remains a challenge.

Furthermore, while the system improves the reasoning capabilities of LLMs using structured knowledge, the methods for evaluating the quality and relevancy of the generated explanations based on graph matching still need refinement. Establishing more robust metrics and evaluation protocols to assess explanation quality in legal contexts could significantly enhance the system's credibility and effectiveness.

\vspace{.5cm}
\subsection{Potential Future Research Directions and Applications}

Moving forward, several research directions could be pursued to address the current limitations and expand the capabilities of the Soft PROLEG system. Exploring advanced algorithms for graph-based reasoning could alleviate computational burdens and improve the efficiency of query handling. Techniques such as graph neural networks or more sophisticated forms of graph algorithms could be adapted to enhance the speed and accuracy of inference in complex legal scenarios.

There is also considerable potential in refining the system's ability to generate and evaluate explanations. Future work could focus on developing machine learning models specifically trained to assess the quality of legal reasoning presented in graphical forms. Employing techniques from the field of explainable AI to improve the generation of explanations based on user feedback adaptively could further increase the system's practicality.

Additionally, expanding the applications of Soft PROLEG to other areas of law beyond the initial scenarios, and possibly other domains like healthcare or finance, where compliance and regulations play significant roles, could showcase the framework's versatility and impact. Cross-domain applications could also help in fine-tuning the system, as diverse challenges would reveal new insights and improvement opportunities.

\section{Conclusion}

The Knowledge Representation Augmented Generation (KRAG) framework, demonstrated by the Soft PROLEG system, constitutes a significant advancement in applying Large Language Models (LLMs) in the legal field, enhancing accuracy, reliability, and interpretability of automated legal reasoning by integrating structured knowledge graphs with LLMs. Soft PROLEG improves handling legal text complexities and fosters user trust with transparent, explainable decisions, making it a strong tool in legal text analysis. Future improvements can include advanced graph processing and automated verification techniques. The application of KRAG extends beyond legal domains to areas like healthcare and finance, indicating a promising future for AI in complex decision-making roles. This step forward with KRAG not only predicts a future of effective human-AI collaboration but actively creates a path towards it, maximizing the potential of AI in enhancing human expertise.

\section*{Appendix}

Below are some additional queries and answers by Soft PROLEG, showing the effectiveness of this KRAG-based system. In actual implementation, the explanation phase of Soft PROLEG also incorporates diagrams that visually represent the reasoning process:

{\fontfamily{qcr}\selectfont
\footnotesize
\begin{mdframed}
\textbf{Query:} Alice sells her car to Bob for \$10,000 on January 5, 2022. However, Bob later discovers that Alice was not the legal owner of the car. Can Bob nullify the contract?

\noindent \textbf{Answer:} In this case, Bob can nullify the contract because the essential condition of Alice's ownership of the car was not met. This absence of ownership is a direct exception to the validity of the contract, thus rendering it voidable at Bob's discretion.
\end{mdframed}
}

This query demonstrates Soft PROLEG's ability to apply legal principles to contractual disputes effectively. The decision hinges on the fundamental requirement of ownership in sales contracts. By recognizing that Alice did not meet this essential condition (i.e., she was not the legal owner), Soft PROLEG logically concludes that the contract between Alice and Bob is voidable. This example effectively showcases the model's capability to pinpoint critical legal conditions and apply exceptions accurately.

{\fontfamily{qcr}\selectfont
\footnotesize
\begin{mdframed}
\textbf{Query:} Linda enters into a contract with a software company to develop a custom software solution for her business. The contract includes a clause that the software must pass specified security tests. After the software is developed, it fails one critical security test. The software company argues that the failure is due to Linda's server specifications, which did not meet their recommended standards—a fact disclosed to Linda at the project's initiation. Linda insists on a breach of contract and demands a refund. Can Linda legally enforce the breach of contract?

\noindent \textbf{Answer:} Linda's case for enforcing a breach of contract hinges on whether the software's failure to meet security tests falls under the software company's responsibility. However, the exception here is that the failure was attributed to Linda's server specifications, which were below the recommended standards and known to Linda. If it is established that the server inadequacy directly caused the security failure, the software company could be exempt from liability for the breach, reducing Linda's grounds for a refund.
\end{mdframed}
}

This example illustrates the complexity of legal arguments in contractual obligations involving technical products. Soft PROLEG skillfully navigates through the layers of responsibility and causality by identifying the exception regarding server specifications. 

We also experimented with the same prompt using GPT-4. This model returned an explanation three times as long with the same conclusion. However, we found its responses tended to include redundant information and lacked the precision needed to stay on-point, which could potentially lead the reader astray.

\section*{Acknowledgments}
This work was supported by AIP challenge funding in relation with JST, AIP Trilateral AI Research, Grant Number JPMJCR20G4, by JSPS KAKENHI Grant Numbers,
JP22H00543, JST, AIP Trilateral AI Research, Grant Number JPMJCR20G4
and the MEXT "R\&D Hub Aimed at Ensuring Transparency and Reliability
of Generative AI Models" project.

\bibliographystyle{kr}
\bibliography{kr-sample}

\end{document}